\documentclass[10pt,twocolumn,letterpaper]{article}

\usepackage{cvpr}              

\usepackage{graphicx}
\usepackage{amsmath}
\usepackage{amssymb}
\usepackage{booktabs}

\usepackage[pagebackref,breaklinks,colorlinks]{hyperref}

\usepackage[capitalize]{cleveref}
\crefname{section}{Sec.}{Secs.}
\Crefname{section}{Section}{Sections}
\Crefname{table}{Table}{Tables}
\crefname{table}{Tab.}{Tabs.}

\def\cvprPaperID{*****} 
\def\confName{CVPR}
\def\confYear{2022}

\begin{document}

\title{1st Place Solution for 5th LSVOS Challenge: \\ Referring Video Object Segmentation}
\author{Zhuoyan Luo$^{1}$\footnotemark[2], Yicheng Xiao$^{1}$\footnotemark[2], Yong Liu$^{1,2}$\footnotemark[2]\footnotemark[3], Yitong Wang$^{2}$, Yansong Tang$^{1}$, Xiu Li$^{1}$, Yujiu Yang$^{1}$\\
\small $^1$Tsinghua Shenzhen International Graduate School, Tsinghua University  $^2$ByteDance Inc. \\
{\tt\small \{luozy23, xiaoyc23, liu-yong20\}@mails.tsinghua.edu.cn} \\
}
\maketitle
\thispagestyle{empty}
\renewcommand{\thefootnote}{\fnsymbol{footnote}}
\footnotetext[2]{Equal Contribution}
\footnotetext[3]{Project lead, done during internships at ByteDance}

\begin{abstract}
The recent transformer-based models have dominated the Referring Video Object Segmentation (RVOS) task due to the superior performance. Most prior works adopt unified DETR framework to generate segmentation masks in query-to-instance manner. In this work, we integrate strengths of that leading RVOS models to build up an effective paradigm. We first obtain binary mask sequences from the RVOS models. To improve the consistency and quality of masks, we propose Two-Stage Multi-Model Fusion strategy. Each stage rationally ensembles RVOS models based on framework design as well as training strategy, and leverages different video object segmentation (VOS) models to enhance mask coherence by object propagation mechanism. Our method achieves $75.7\%$ $\mathcal{J} \& \mathcal{F}$ on Ref-Youtube-VOS validation set and 
$70\%$ $\mathcal{J} \& \mathcal{F}$ on test set, which ranks 1st place on 5th Large-scale Video Object Segmentation Challenge (ICCV 2023) track 3. Code is available at \href{https://github.com/RobertLuo1/iccv2023_RVOS_Challenge}{https://github.com/RobertLuo1/iccv2023\_RVOS\_Challenge}.
\end{abstract}
\vspace{-10pt}
\section{Introduction}
\label{sec:intro}
Referring Video Object Segmentation aims to segment and track the target object referred by the given text description in a video.
This emerging field has garnered attention due to its potential applications in video editing and human-robot interaction. 

The critical challenge in RVOS lies in the pixel-level alignment between different modalities and time steps, primarily due to the varied nature of video content and unrestricted language expression. 
Most early approaches~\cite{refvos,rel1,rel2} adopt multi-stage and complex pipelines that take the bottom-up or top-down paradigms to segment each frame separately, while recent works MTTR~\cite{mttr}, Referformer~\cite{referformer} propose to unify cross-modal interaction with pixel-level understanding into transformer structure.
For example, 2022 first winner~\cite{last_champ} simply employs fine-tuned Referformer as backbone to generate a series of high quality masks.
However, these methods may lose the perception of target objects for language descriptions expressing temporal variations of objects due to the lack of video-level multi-modal understanding.
To address this issue, SOC~\cite{soc}, MUTR~\cite{mutr} efficiently aggregate inter and intra-frame information.
Meanwhile, UNINEXT~\cite{UNINEXT} proposes a unified prompt-guided formulation for universal instance perception, reuniting previously fragmented instance-level sub-tasks into a whole and achieve good performance for the RVOS task.

\begin{figure}[t]
\begin{center}
\includegraphics[width=85mm]{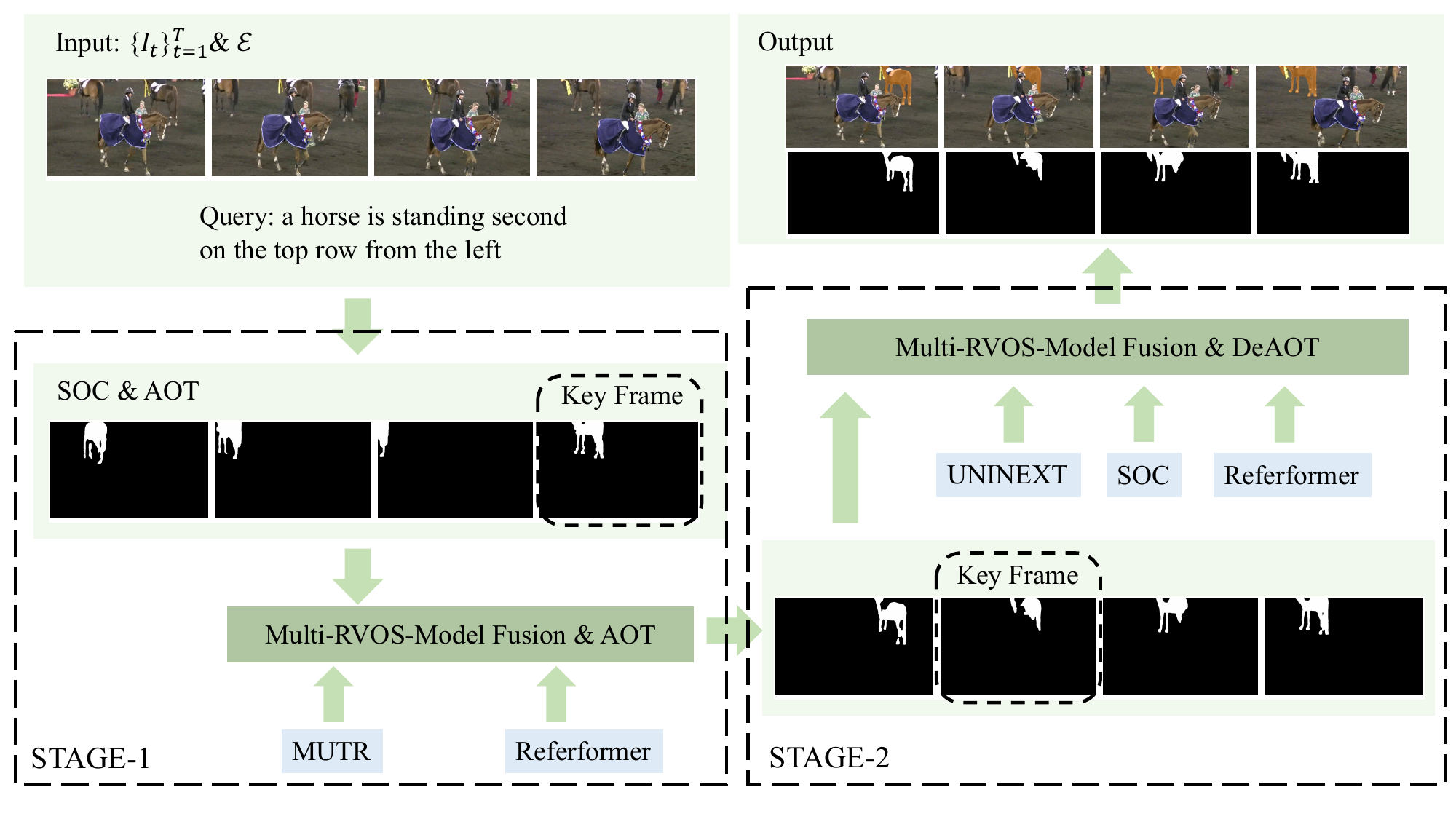}
\end{center}
\vspace{-10pt}
\caption{The overall architecture of our method.}
\label{fig:fig1}
\end{figure}
In our work, we incorporate benefits of the previous mainstream works to provide an effective paradigm. 
By utilizing the model ensemble strategy as well as semi-supervised VOS approaches as post-process to enhance the masks quality in each stage, we develop a Two-Stage Multi-model Fusion strategy.
Specifically, we select AOT~\cite{aot} to preliminary improve the masks quality in the first stage, but with the increase of propagation layers and number of RVOS models that are processed by that, it will inevitably lead to loss of information unrelated to the object, which may weaken the effect of the consequent model fusion.
Therefore, on the basis of the high quality mask sequences from the first stage, we further exploit the potential of multi-model fusion by utilizing DeAOT~\cite{deaot} in the second stage.

The final leaderboard shows that our method ranks 1st place in the 5th Large-scale Video Object Segmentation Challenge (ICCV 2023): Referring Video Object Segmentation track.


\section{Related Work}
\label{sec:relate}

\paragraph{Semi-supervised Video Object Segmentation} 
The objective of Semi-supervised Video Object Segmentation (VOS) is to achieve accurate object segmentation throughout the entire video sequence by utilizing provided one (generally at the first frame) or more mask annotation.
The works~\cite{semi1,semi2} focus on improving run-time efficiency and matching feature correlation between the target and other potential objects in the sequence to enhance the object tracking process.
In addition, FEELVOS~\cite{feelvos} 
extends the pixel-level matching mechanism by additionally doing local matching with the previous frame.
STM~\cite{stm} leverages a memory network to store past-frame predictions and apply attention mechanism to propagate the mask information.
Recently, AOT~\cite{aot} introduces hierarchical propagation into VOS and employs an identification mechanism to associate multiple targets.
Furthermore, DEAOT~\cite{deaot} decouples object-agnostic and object-specific features in hierarchical propagation.
In this work, we include two methods mentioned above as post-process.

\paragraph{Referring Video Object Segmentation.} 
Referring Video Object Segmentation (RVOS) is first proposed by Gavrilyuk \textit{et.al.} \cite{GavrilyukGLS18}, which aims to generate a series of binary segmentation masks of the instance referred by the natural language description across a video clip.
URVOS~\cite{urvos} introduces a large-scale RVOS benchmark and a unified framework that leverages attention mechanisms and mask propagation with a semi-supervised VOS method.
Similar to URVOS~\cite{urvos}, some approaches~\cite{refvos,rel1} process each frame of the video clip separately through an image-level model.
Meanwhile, compared to~\cite{rvos1, rvos2} rely on complicated pipelines, MTTR~\cite{mttr} and Referformer~\cite{referformer} first adopt end-to-end framework modeling the task as the a sequence prediction problem, which greatly simplifies the pipeline.
Currently, SOC~\cite{soc} and MUTR~\cite{mutr} achieve excellent performance by efficiently aggregating intra and inter-frame information.
What's more, UNINEXT~\cite{UNINEXT} reformulates diverse instance perception tasks into a unified object discovery and retrieval paradigm.
In this work, we combine the advantages of the above methods to obtain high-quality mask sequences.

\begin{figure*}[t]
\begin{center}
\includegraphics[width=\linewidth]{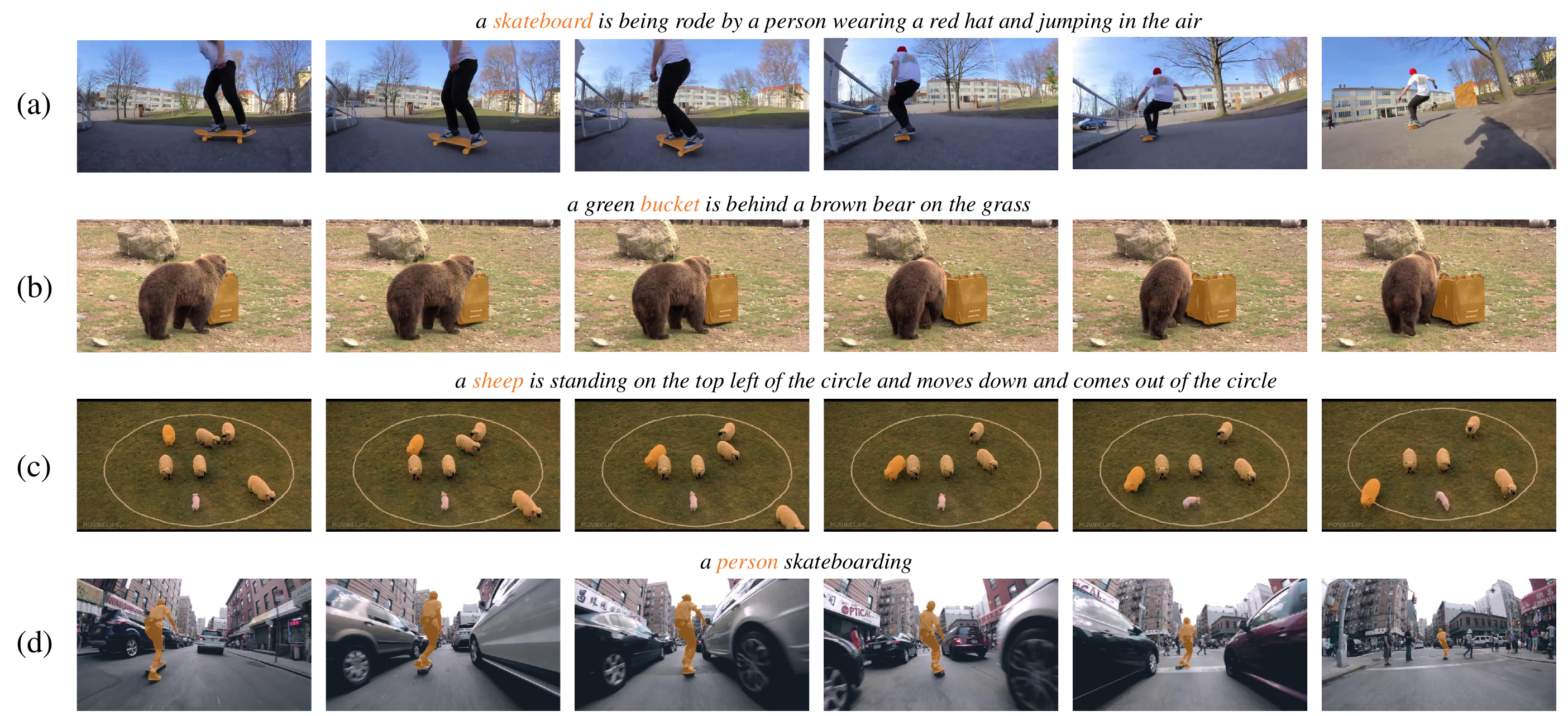}
\end{center}
\vspace{-10pt}
\caption{Visualization results on Ref-Youtube-VOS.}
\label{fig:fig2}
\end{figure*}

\section{Method}
Given $T$ frames of video clip $\mathcal{I} = \{I_t\}_{t=1}^T$, where $I_t \in \mathbb{R}^{3 \times H_0 \times W_0}$ and a referring text expression $\mathcal{E} = \{e_i\}_{i=1}^L$, where $e_i$ denotes the i-th word in the text. 
RVOS task is to generate a series of binary segmentation masks $\mathcal{S} = \{s_t\}_{t=1}^T$, $s_t \in \mathbb{R}^{1\times H_0 \times W_0}$ of the referred object. 

\subsection{Backbone}
We adopt SOC~\cite{soc}, MUTR~\cite{mutr}, Referformer~\cite{referformer} and UNINEXT~\cite{UNINEXT}, the current prevalent RVOS models, as our backbones to  respectively generate binary segmentation masks $\mathcal{S} = \{s_t\}_{t=1}^T$. 
\begin{equation}
    \mathcal{S}^{n} = \mathcal{F}^{n}\left( \mathcal{I}, \mathcal{E}\right) \quad n \in \{soc, mutr, ref, uninext\},
\end{equation}
where $\mathcal{F}^{n}$ indicates the corresponding backbone. 
We train SOC jointly on RefCOCO and Ref-Youtube-VOS datasets, while we directly use checkpoints with the highest performance without training for other models. 
\subsection{Post-process}
The video object segmentation has been proved to improve the segmentation mask consistency by object propagation mechanism. Specifically, \cite{last_champ} adopts AOT~\cite{aot} as post-process to enhance the quality of mask results generated by RVOS models, which brings a clear improvement in accuracy. The general procedure are first selecting the key-frame index of mask sequences probability $\mathcal{P}$ from RVOS model, then using VOS model to perform forward and backward propagation. It can be formulated as:
\begin{equation}
\begin{gathered}
    \mathcal{K}_{index} = argmax(\mathcal{P}), \\
    \mathcal{M}^n = \left[\mathcal{G}\left(\{s^n_{i}\}_{i=K_{index}}^{0}\right), \mathcal{G}\left(\{s^n_{j}\}_{j=K_{index}}^{T}\right)\right], \\
\end{gathered}
\end{equation}
where $\mathcal{G}$ denotes the VOS model for post-process. 

In our experiment, we find that although AOT can facilitate the temporal quality of mask results, the benefit decreases when conducting the Stage \uppercase\expandafter{\romannumeral2} fusions (which is elaborate on~\cref{sec:ensemble}). It is hypothesized that AOT potentially lead to loss of object-agnostic visual information in deep propagation layer. Consequently, the advantage of ensemble is degraded due to the aggregated object information loss from different RVOS models that are post-processed by the the same VOS model. Intuitively, we propose to use two VOS models for post-processing in different stages to alleviate the problem. 

\subsection{Two-Stage Multi-model Fusion}
\label{sec:ensemble}
We find that models with different frameworks process the object referred by the textual description in different perspectives. SOC unifies temporal modeling and cross modal alignment to achieve video-level understanding, which comprehends expressions containing temporal variations well. Due to the object discovery and retrieval paradigm, UNINEXT has strong ability of localizing and tracking same objects referred by different textual descriptions. MUTR introduces temporal transformer to facilitate objects interaction across frames, improving consistency of masks. In order to make full use of advantages of different frameworks, we propose two-stage multi-model fusion strategy. Similar to ~\cite{last_champ}, we fuse the masks predicted by different referring expressions that describe the same target from different models, which is formulated as:
\begin{equation}
    \begin{gathered}
        \hat{y} = \sum_{n}^{N}\sum^{Q}_{q}\mathcal{M}^n_q, \\    
        \hat{y'}_i = \begin{cases}0 & \hat{y}_i<thr \\ 1 & \hat{y}_i>=thr\end{cases},
    \end{gathered}
\end{equation}
where $Q$ denotes the number of different textual descriptions referred to the same object and $N$ indicates the models. $i \in \{1,2, \dots, HW\}$ where $H$, $W$ are the width and height of the mask respectively.  

\textbf{Stage \uppercase\expandafter{\romannumeral1}} Referformer treats the task as sequence prediction problem and perform cross modal interaction in each frame. 
Its simple framework could serve as a baseline to segment the referred object but easily fails to capture the temporal variation of object across frames. 
We believe that SOC and MUTR can explicitly increase the inter-frame interaction, which is a reasonable compensation for Referformer. 
Therefore, in the first stage, we fuse three models and use AOT as post-process to enhance the mask quality. For clarity, the fused model is denoted as SMR.

\textbf{Stage \uppercase\expandafter{\romannumeral2}} UNINEXT is jointly trained with prevalent datasets of $10$ instance perception tasks. 
It is capable of perceiving diverse objects referred by different descriptions, thanks to static object queries which absorb rich information from data in different domain. 
Although it achieve high performance with VIT-Huge backbone~\cite{vith} by feeding large scale of data, the lack of global view of object may cause the inconsistency when generating masks across frames. 
Therefore, we solve this problem by two-fold. (1) Employ DeAOT to propagate the object information from the key frame to another. (2) Ensemble with the SMR fused model to integrate information from inter-frame interaction.

\begin{table}
\begin{center}
\begin{tabular}{c||c}
\toprule[1.1pt]
\textbf{Model} & {$\mathcal{J}$ \& $\mathcal{F}$} $\uparrow$ \\ 
\hline\hline
SOC & 67.5 \\
\hline
+AOT & 69.5 \textcolor{red}{(+2.0)} \\
+Multi-model Fusion (Stage \uppercase\expandafter{\romannumeral1}) \& AOT & 72.4 \textcolor{red}{(+2.9)} \\
+Multi-model Fusion (Stage \uppercase\expandafter{\romannumeral2}) \& DeAOT & 75.7 \textcolor{red}{(+3.3)} \\
\hline
\end{tabular}
\end{center}
\vspace{-10pt}
\caption{Ablation study of each module on our model’s performance on \textbf{validation set}.}
\label{table:ablation}
\end{table}

\section{Experiment}

\subsection{Dataset and Metrics} 

\paragraph{Datasets.} We evaluate our model on Ref-Youtube-VOS dataset of \textit{2023 Referring Youtube-VOS challenge}. It contains 3,978 high-resolution YouTube videos with about 15K language expressions. These videos are divided into 3,471 training videos, 202 validation videos and 305 test videos.

\paragraph{Metrics.} we adopt standard evaluation metrics: region similarity ($\mathcal{J}$), contour accuracy ($\mathcal{F}$) and
their average value ($\mathcal{J \& \mathcal{F}}$) on Ref-Youtube-VOS.

\subsection{Training Detail} 
We train SOC with pretrained Video Swin Transformer and RoBERTa as the encoder for 30 epochs. The model is optimized by Adam optimizer with the initial learning rate of 1e-4. During training, we apply RandomResize and Horizontal Flip for data augmentation.
Specifically, all frames are downsampled to 360×640. In post-process, we follow~\cite{last_champ} retrain DeAOT network with Swin-L backbone using default parameters in~\cite{deaot}.


\subsection{Main Results}
Our method achieves $70\%$ $\mathcal{J} \& \mathcal{F}$ on test set which outperforms the next team by $4\%$ $\mathcal{J} \& \mathcal{F}$ and rank 1st place in Large-scale Video Object Segmentation Challenge (ICCV 2023): Referring Video Object Segmentation track.

\subsection{Ablation Study} 
To validate the effectiveness of each module, we conduct simple ablation studies. As we mention above that we use SOC~\cite{soc}, MUTR~\cite{mutr}, Referformer~\cite{referformer} and UNINEXT~\cite{UNINEXT} as RVOS models to generate mask results for post-processing and fusion. It is noted that SOC is the main model that are included in two stages, and for simplicity, we set it as the baseline. As shown in \cref{table:ablation}, the preliminary model fusion and post-process with AOT in stage \uppercase\expandafter{\romannumeral1},  brings an improvement of $2.9\%$ $\mathcal{J} \& \mathcal{F}$. While the fused model achieve $75.7\%$ $\mathcal{J} \& \mathcal{F}$ with the second stage model ensemble, demonstrating the rationality and significance of our proposed two-stage multi models fusion.

\subsection{Qualitative Results}
\cref{fig:fig2} shows the prediction of our method for complex scenarios segmentation, \textit{i.e.}, similar appearance, occlusion and large variations.
It can be seen that our method precisely segments the referred object.

{\small
\bibliographystyle{ieee_fullname}
\bibliography{egbib}
}

\end{document}